\documentclass[10pt,twocolumn,letterpaper]{article}

\usepackage[pagenumbers]{cvpr} % To force page numbers, e.g. for an arXiv version

\newcommand{\TODO}[1]{\textbf{\color{red}[TODO: #1]}}
\renewcommand{\TODO}[1]{}

\usepackage{booktabs}
\usepackage{subcaption}
\usepackage{graphicx}
\usepackage{amsmath}
\usepackage{xspace}
\usepackage[accsupp]{axessibility}

\definecolor{cvprblue}{rgb}{0.21,0.49,0.74}
\usepackage[pagebackref,breaklinks,colorlinks,allcolors=cvprblue]{hyperref}

\def\paperID{MAI-20} % *** Enter the Paper ID here
\def\confName{CVPR}
\def\confYear{2026}

\newcommand{\method}{Slimmable ConvNeXt\xspace}

\title{Slimmable ConvNeXt: Width-Adaptive Inference \\for Efficient Multi-Device Deployment}

\author{Janek Haberer\\
Kiel University, Germany\\
{\tt\small janek.haberer@cs.uni-kiel.de}
\and
Jon Eike Wilhelm\\
Kiel University, Germany\\
{\tt\small }
\and
Olaf Landsiedel\\
Hamburg University of Technology (TUHH), Germany\\
Kiel University, Germany\\
UNU-INWEH, Germany\\
{\tt\small olaf.landsiedel@tuhh.de}
}

\begin{document}
\maketitle
\begin{abstract}

Deploying vision models across devices with varying resource constraints, or even on a single device where available compute fluctuates due to battery state, thermal throttling, or latency deadlines, typically requires training and maintaining separate models.
Width-adaptive inference addresses this by training a single set of shared weights containing multiple nested subnetworks of increasing capacity, but prior CNN-based approaches required switchable batch normalization, while recent scalable methods have focused on Vision Transformers.
We present \method, which shows that ConvNeXt's modern design, specifically LayerNorm and inverted bottlenecks, makes it particularly suited for channel-width slimming, eliminating the normalization overhead of classical slimmable networks and producing a simpler training pipeline than both prior CNN and ViT approaches.
On ImageNet-1k, \method-T with 3 subnetworks achieves 80.8\% top-1 accuracy at 4.5~GMACs and 77.4\% at 1.2~GMACs, trained from scratch for 600 epochs. At comparable compute, this exceeds HydraViT's 6-head subnetwork (78.4\% at 4.6~GMACs) by 2.4 percentage points and its 3-head configuration (73.0\% at 1.3~GMACs) by 4.4 percentage points, while also outperforming MatFormer-S (78.6\%) and SortedNet-S (78.2\%) at the same GMACs. Scaling to \method-B further improves maximum accuracy to 82.8\% at 15.35~GMACs.

\end{abstract}

% Hero figure at bottom of page 1
\begin{figure}[!b]
  \centering
  \includegraphics[width=\linewidth]{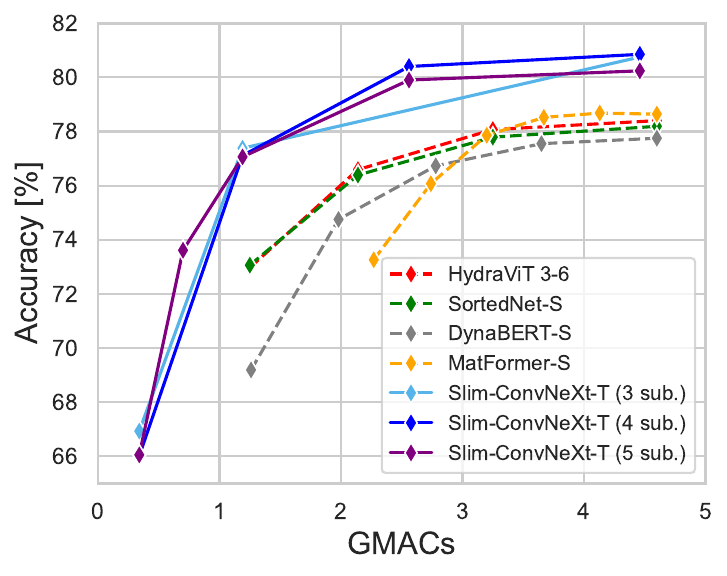}
  \caption{\textbf{ConvNeXt-T variants vs.\ ViT-S-based baselines on ImageNet-1k.} \method-T with 3, 4, and 5 subnetworks compared to HydraViT, SortedNet, DynaBERT, and MatFormer in the low-compute regime. All ConvNeXt-T variants achieve a higher accuracy at significantly lower GMACs.}
  \label{fig:hero}
\end{figure}

\section{Introduction}
\label{sec:intro}

Deploying vision models across devices with varying resource constraints remains a key challenge~\cite{cai_once-for-all_2020,tan_efficientnet_2019,afifi_ml_networks_2024}.
Practitioners typically train and store separate models for each target platform, for example a large model for cloud inference and a smaller one for mobile or edge devices.
This is costly and becomes impractical when hardware availability changes or when the same application must serve a heterogeneous fleet of devices.
Even on a single device, available compute may fluctuate at runtime due to battery state, thermal throttling, or latency deadlines~\cite{fang_nestdnn_2018,huang_multi-scale_2018}.
Scalable architectures address both scenarios by training a single set of weights containing multiple nested subnetworks of increasing capacity, allowing dynamic selection of the computational budget at inference time.

Channel-width slimming for CNNs was pioneered by Slimmable Networks~\cite{yu_slimmable_2018} and US-Nets~\cite{yu_universally_2019}, which demonstrated that a single ResNet or MobileNet can operate at multiple channel widths with shared weights, but require switchable batch normalization and specialized training techniques.
More recently, scalable inference has shifted to Vision Transformers (ViTs)~\cite{dosovitskiy_image_2021}, such as HydraViT~\cite{haberer_hydravit_2024}, which extracts subnetworks by adding and dropping attention heads, while MatFormer~\cite{devvrit2024matformer} flexes the MLP hidden dimension for a more limited scalability range.

We observe that ConvNeXt~\cite{liu_convnet_2022}, a modernized CNN with depthwise convolutions, inverted bottlenecks, and LayerNorm, is particularly well-suited for width slimming compared to both classical CNNs and Vision Transformers.
LayerNorm eliminates switchable batch normalization entirely, the inverted bottleneck structure concentrates parameters in pointwise layers that are straightforward to slice, and the absence of self-attention avoids the associated overhead while ensuring broad hardware compatibility from cloud GPUs to mobile devices.
Combining these properties with width-adaptive slimming yields an elastic model that can scale its compute on the fly.
Yet, to our knowledge, no prior work has applied width-adaptive slimming to ConvNeXt.

\begin{figure}[t]
\centering
\includegraphics[width=\linewidth]{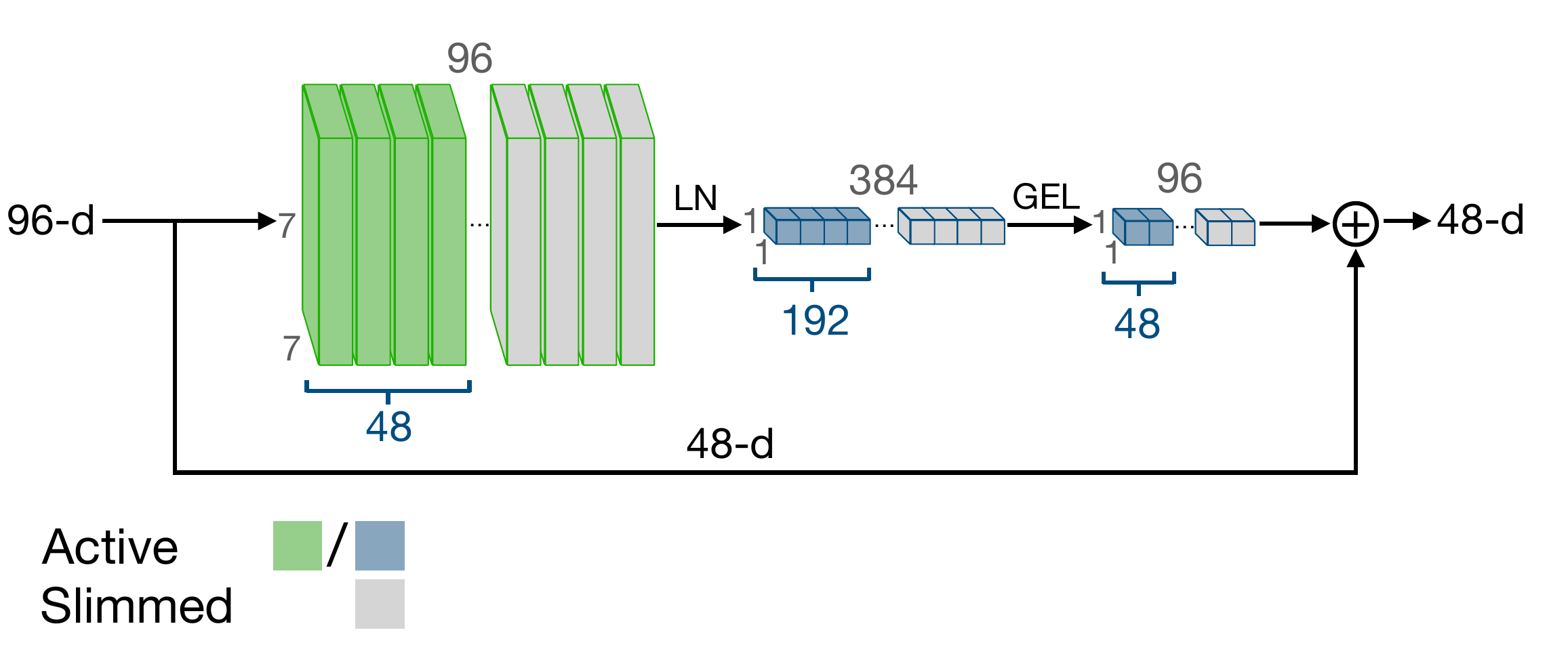}
\caption{A ConvNeXt block slimmed to $p{=}0.5$. All weight tensors are sliced to retain the first $\lfloor p \cdot C \rfloor$ channels, including the $4{\times}$ expanded intermediate layer. The residual is zero-padded to match the input dimension.}
\label{fig:slim_block}
\end{figure}

In this paper, we propose \method, which assigns each ConvNeXt block a slimming ratio $p \in (0, 1]$ controlling the fraction of active channels (see Figure~\ref{fig:slim_block}).
During training, we alternate between width configurations, jointly optimizing all subnetworks with shared weights without requiring switchable normalization, knowledge distillation, or the sandwich training rule.
We further investigate AutoSlim~\cite{yu_autoslim_2019} for non-uniform per-block widths and scale our approach from ConvNeXt-T to ConvNeXt-S and ConvNeXt-B on ImageNet-1k~\cite{deng_imagenet_2009}.

Our contributions are as follows:
\begin{enumerate}
\itemsep0em
    \item We show that ConvNeXt's design choices, in particular LayerNorm and inverted bottlenecks, make it an ideal architecture for width-adaptive slimming, eliminating the overhead required by prior CNN approaches~\cite{yu_slimmable_2018,yu_universally_2019}.
    \item On ImageNet-1k, \method-T with 3 subnetworks achieves 80.8\% at 4.5~GMACs and 77.4\% at 1.2~GMACs, trained from scratch for 600 epochs. This exceeds HydraViT's~\cite{haberer_hydravit_2024} 6-head subnetwork (78.4\%) by 2.4 percentage points, while also outperforming MatFormer-S~\cite{devvrit2024matformer} and SortedNet-S~\cite{valipour2023sortednet} at the same GMACs (see Figure~\ref{fig:hero}). Scaling to \method-B further improves maximum accuracy to 82.8\% at 15.35~GMACs.
    \item We evaluate AutoSlim for per-block width optimization and analyze trade-offs between subnetwork count, model scale, and training duration across three model sizes.
\end{enumerate}

\section{Related Work}
\label{sec:related}

\begin{figure*}[t]
\centering
\includegraphics[width=\linewidth]{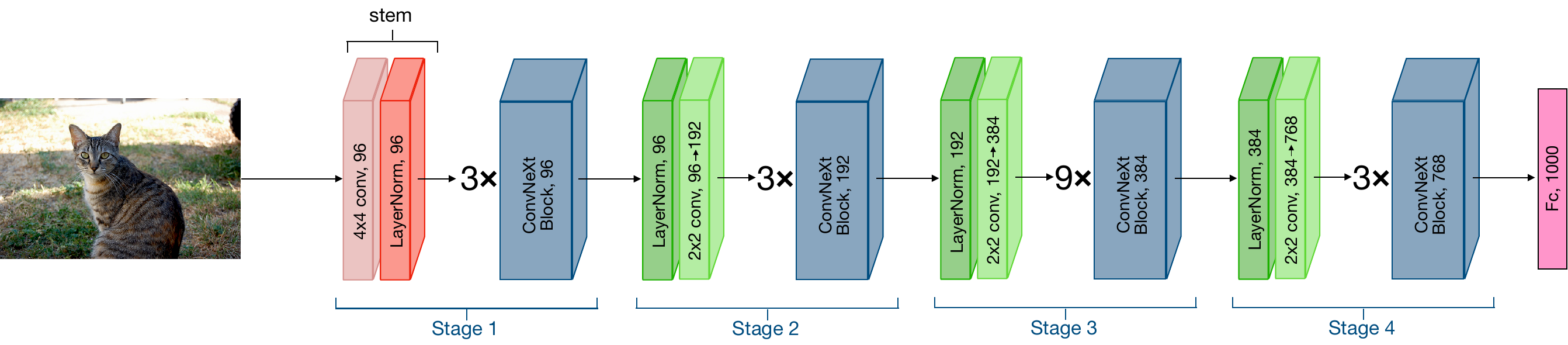}
\caption{The ConvNeXt-Tiny architecture with 4 stages and 18 blocks total. Downsampling layers between stages halve spatial resolution and double channel count.}
\label{fig:convnext_arch}
\end{figure*}

Training a single set of shared weights containing multiple nested subnetworks of increasing capacity has a long history in CNNs.
Slimmable Networks~\cite{yu_slimmable_2018} train one model at multiple predefined channel widths using \emph{switchable batch normalization} to maintain separate running statistics per width.
US-Nets~\cite{yu_universally_2019} extend this to arbitrary widths but require a \emph{sandwich rule} and \emph{inplace knowledge distillation} to stabilize training.
NestDNN~\cite{fang_nestdnn_2018} creates variants by iteratively pruning filters for resource-aware on-device inference.
AutoSlim~\cite{yu_autoslim_2019} complements these with a greedy per-layer channel search for non-uniform width configurations under a compute budget.
Once-for-All~\cite{cai_once-for-all_2020} supports diverse architectural configurations (depth, width, kernel size, resolution) and uses neural architecture search to extract specialized subnetworks.
These methods establish shared-weight, multi-width inference for CNNs but most rely on BatchNorm-based networks, necessitating switchable normalization.

The concept of learning ordered representations is closely related.
Nested Dropout~\cite{rippel_learning_2014} trains representations whose dimensions are ordered by importance by randomly dropping ordered sets of hidden units.
Matryoshka Representation Learning~\cite{kusupati_matryoshka_2024} trains embeddings at multiple granularities, enabling truncation to different lengths depending on the compute budget.
Both share the principle that training with structured subsets induces graceful degradation when capacity is reduced.

For Vision Transformers, HydraViT~\cite{haberer_hydravit_2024} exploits the nested structure of ViT~\cite{dosovitskiy_image_2021} configurations: subnetworks are extracted by taking the first $k$ attention heads and slicing all embedding and weight matrices accordingly.
Using stochastic tail-drop training, HydraViT induces up to 10 subnetworks within a single ViT-B, achieving performance comparable to individually trained DeiT~\cite{touvron2021training} models.
MatFormer~\cite{devvrit2024matformer} makes only the MLP hidden dimension flexible via Matryoshka-style nesting, limiting its scalability range to approximately half of the full compute.
DynaBERT~\cite{hou_dynabert_2020} adjusts both attention heads and MLP width but keeps the full embedding dimension between blocks and relies on knowledge distillation~\cite{hinton_distilling_2015}.
SortedNet~\cite{valipour2023sortednet} generalizes nesting across depth and width but keeps the number of heads fixed, introducing scaling inconsistencies as the embedding dimension shrinks.
ThinkingViT~\cite{hojjat2025thinkingvit} builds on top of nested ViTs and makes them input-adaptive, such that the network itself decides how much compute it spends on a specific input.

Efficient convolutional architectures have been designed with deployment constraints in mind.
MobileNetV2~\cite{sandler_mobilenetv2_2018} introduced inverted bottlenecks with depthwise separable convolutions, a design pattern ConvNeXt later adopted.
EfficientNet~\cite{tan_efficientnet_2019} proposed compound scaling of depth, width, and resolution.
ConvNeXt~\cite{liu_convnet_2022} modernized ResNet~\cite{he2016deep} by adopting Transformer design choices: a patchify stem, depthwise convolutions, inverted bottlenecks, LayerNorm, and GELU activations, bringing pure CNN accuracy from 76.1\% (ResNet-50) to 82.1\% (ConvNeXt-T), surpassing Swin Transformer~\cite{liu_swin_2022} at comparable compute.

Despite this, no prior work has applied width-adaptive slimming to ConvNeXt.
We show that its LayerNorm eliminates switchable batch normalization~\cite{yu_slimmable_2018,yu_universally_2019} and its inverted bottlenecks concentrate parameters in pointwise layers that are straightforward to slice, yielding a simpler pipeline than both prior CNN and ViT approaches while achieving higher accuracy than state-of-the-art baselines at nearly every operating point.

\section{Method}
\label{sec:method}

We first review the ConvNeXt architecture, then describe channel-width slimming, the joint training procedure, and the optional AutoSlim search for non-uniform per-block widths.

\subsection{Design Overview}
\label{sec:overview}

Each ConvNeXt block is assigned a slimming ratio $p \in (0,1]$ that determines what fraction of its channels remain active. All weight tensors (depthwise convolution, pointwise convolutions, and LayerNorm) are sliced to the first $\lfloor p \cdot C \rfloor$ channels, and the residual is zero-padded to preserve dimensions. During training, a set of width configurations (p-lists) defines the available subnetworks. Each batch randomly selects one configuration and performs a standard forward-backward pass through the corresponding subnetwork. Because ConvNeXt uses LayerNorm rather than Batch Normalization, no switchable normalization is needed, and the model handles all widths with a single set of normalization parameters. At inference time, any trained width can be selected to match the device's compute budget or deadline, without reloading weights. Optionally, AutoSlim refines the uniform per-block widths into non-uniform assignments that allocate more capacity to important blocks.

\subsection{ConvNeXt Preliminaries}
\label{sec:convnext_prelim}

ConvNeXt~\cite{liu_convnet_2022} is organized into four stages, each containing a sequence of identical blocks. A block applies a $7{\times}7$ depthwise convolution, followed by LayerNorm, a pointwise ($1{\times}1$) convolution that expands the channel dimension by a factor of four, a GELU activation, and a second pointwise convolution that projects back to the original dimension. A residual connection~\cite{he2016deep} adds the block input to its output. This inverted bottleneck design, where the intermediate representation is wider than the input and output, mirrors the structure of MobileNetV2~\cite{sandler_mobilenetv2_2018} and modern Transformer MLP layers. Between stages, a downsampling layer applies LayerNorm followed by a $2{\times}2$ convolution with stride~2, halving the spatial resolution and doubling the number of channels.

\begin{table}[t]
\caption{ConvNeXt model configurations used in this work.}
\label{tab:convnext_sizes}
\centering
\setlength{\tabcolsep}{4pt}
\footnotesize
\begin{tabular}{@{}lcccc@{}}
\toprule
Model & Depths & Dims & Params & GMACs \\
\midrule
ConvNeXt-T & (3,3,9,3) & (96,192,384,768) & 28.6M & 4.5 \\
ConvNeXt-S & (3,3,27,3) & (96,192,384,768) & 50.2M & 8.7 \\
ConvNeXt-B & (3,3,27,3) & (128,256,512,1024) & 88.6M & 15.4 \\
\bottomrule
\end{tabular}
\end{table}

Figure~\ref{fig:convnext_arch} shows the overall architecture. The model begins with a patchify stem, a $4{\times}4$ convolution with stride 4 followed by LayerNorm, which produces feature maps at $\frac{1}{4}$ of the input resolution. Four stages then follow, each consisting of a downsampling layer (LayerNorm + $2{\times}2$ stride-2 convolution) and a sequence of ConvNeXt blocks. The final classification head applies global average pooling and a linear layer.

We consider three model sizes (Table~\ref{tab:convnext_sizes}): ConvNeXt-T (18 blocks), ConvNeXt-S (36 blocks, deeper third stage), and ConvNeXt-B (36 blocks, wider channels). Since the computational cost of each block scales quadratically with the channel dimension (the two pointwise convolutions dominate), slimming to ratio $p$ reduces a block's cost roughly by a factor of $p^2$.

\subsection{Channel-Width Slimming}
\label{sec:slimming}

To create subnetworks at reduced compute, we assign each of the $N$ ConvNeXt blocks a slimming ratio $p \in (0, 1]$ that controls the fraction of active channels. Given a block with $C$ channels, only the first $\lfloor p \cdot C \rfloor$ channels are retained. All weight and bias tensors within the block, including the depthwise convolution, both pointwise convolutions, and LayerNorm parameters, are sliced to match the reduced dimension. The intermediate expansion layer uses $4 \cdot \lfloor p \cdot C \rfloor$ neurons, preserving the original $4{\times}$ expansion ratio. Figure~\ref{fig:slim_block} illustrates a ConvNeXt block slimmed to 50\%.

When a slimmed block produces fewer channels than its input, the residual connection requires matching dimensions. We address this by zero-padding the block output to restore the original channel count before adding the residual. This avoids introducing additional parameters, unlike a learned projection, though it means unused channels lose their identity signal from the skip connection.

This design differs from elastic ViT methods in an important way. All ViT-based approaches face structural constraints imposed by the attention mechanism: HydraViT~\cite{haberer_hydravit_2024} extracts subnetworks by taking the first $k$ out of $H$ attention heads, restricting granularity to multiples of the head dimension $E/H$, where $E$ is the size of the embeddings. DynaBERT~\cite{hou_dynabert_2020} shares this head-dropping constraint and additionally retains the full embedding dimension between blocks, increasing GMACs at small configurations. SortedNet~\cite{valipour2023sortednet} keeps all $H$ heads fixed and only varies the embedding dimension, introducing scaling inconsistencies in attention. MatFormer~\cite{devvrit2024matformer} only flexes the MLP hidden dimension, limiting its scalability range. Because ConvNeXt has no attention mechanism, our channel slicing operates at arbitrary ratios without any such structural constraints, providing finer control over the compute-accuracy trade-off. In principle, any $p \in (0, 1]$ is valid, though in practice we use a discrete set of ratios during training.

\subsection{Joint Training with Alternating Widths}
\label{sec:training}

We define a set of \emph{p-lists}, where each p-list specifies a slimming ratio for every block in the network. For a model with $K$ subnetworks, we use $K$ p-lists. In the simplest case, all blocks share the same ratio within a p-list. For example, training with three subnetworks uses the p-lists $\{0.25, \dots, 0.25\}$, $\{0.5, \dots, 0.5\}$, and $\{1.0, \dots, 1.0\}$, each of length $N$ (18 for Tiny, 36 for Small and Base).

During training, we randomly select one p-list per batch and perform a standard forward and backward pass through the corresponding subnetwork. Formally, at each iteration we sample $k \sim \mathcal{U}(\{1, \dots, K\})$ and optimize:
\begin{equation}
\min_{\theta_k} \sum_{i=1}^{B} \mathcal{L}\bigl(f_{\theta_k}(x_i),\, y_i\bigr),
\label{eq:stochastic_training}
\end{equation}
where $\theta_k \subseteq \theta_K$ denotes the parameters of the $k$-th subnetwork and $B$ is the batch size. All subnetworks share the same weights: a subnetwork at ratio $p < 1$ uses a strict subset of the parameters used at $p = 1$, similar to nested dropout~\cite{rippel_learning_2014}. This stochastic alternation is analogous to the tail-drop training in HydraViT, but applied to channel dimensions rather than attention heads.

An important simplification arises from ConvNeXt's use of LayerNorm instead of Batch Normalization. Slimmable Networks~\cite{yu_slimmable_2018} require switchable batch normalization layers to maintain separate running statistics for each width, adding complexity and parameters. LayerNorm computes statistics per sample and per channel slice, so it adapts naturally to the current channel count without requiring any additional normalization layers.

We follow the training recipe of~\citet{liu_convnet_2022}: AdamW~\cite{loshchilov_decoupled_2019} with cosine learning rate decay, stochastic depth~\cite{huang_deep_2016}, and standard data augmentation including RandAugment~\cite{cubuk_randaugment_2020}, Mixup~\cite{zhang_mixup_2018}, CutMix~\cite{yun_cutmix_2019}, and Random Erasing~\cite{zhong_random_2020}. Full training details are given in Section~\ref{sec:setup}.

\subsection{AutoSlim: Per-Block Width Search}
\label{sec:autoslim}

The uniform p-lists described above assign the same slimming ratio to every block. However, not all blocks contribute equally to accuracy: early blocks extract low-level features and late blocks produce the final representation, while intermediate blocks may be more redundant~\cite{cai_flextron_2024, chavan_vit-slim_2022}. AutoSlim~\cite{yu_autoslim_2019} searches for non-uniform configurations that allocate more capacity to important blocks while slimming less important ones further.

Given a target average ratio $\bar{p}$, the search proceeds greedily: starting from a p-list where all entries are $1.0$, we reduce one entry by a step size of $0.1$ at each iteration. For each of the $N$ blocks, we temporarily reduce its ratio and evaluate accuracy on the validation set. We then commit the reduction that causes the smallest accuracy drop. This process repeats until the average ratio reaches~$\bar{p}$, yielding a non-uniform p-list that reflects the relative importance of each block.

As a concrete example, consider ConvNeXt-T with $N{=}18$ blocks and a target of $\bar{p}{=}0.5$. The search starts from a p-list of all $1.0$s and requires $\lceil(1.0 - 0.5) \times 18 / 0.1\rceil = 90$ iterations to bring the average down to $0.5$, evaluating all 18 blocks at each step. The resulting non-uniform p-list typically preserves wider early and late blocks (Stages 1 and 4) while slimming the middle stages (Stages 2 and 3) more aggressively.

Following the original AutoSlim procedure~\cite{yu_autoslim_2019}, we first train the model with uniform p-lists for a warmup phase (100 epochs for 600-epoch training, 50 for 300-epoch training), then perform the greedy search on this checkpoint. Training then continues from the same checkpoint using only the discovered non-uniform p-lists for the remaining epochs.

\section{Evaluation}
\label{sec:experiments}

\begin{table*}[t]
\caption{\textbf{ImageNet-1k top-1 accuracy (\%) and GMACs for \method-S/B and ViT-B baselines.} All \method models are trained from scratch for 600 epochs. All baselines are trained for 300 (or 800) epochs from a 300-epoch pretrained DeiT-tiny checkpoint, resulting in a total of 600 (or 1100) epochs of training.}
\label{tab:main_results}
\centering
% \small
\begin{tabular}{@{}lcccccccc@{}}
\toprule
& \multicolumn{2}{c}{$p{=}0.25$} & \multicolumn{2}{c}{$p{=}0.5$} & \multicolumn{2}{c}{$p{=}0.75$} & \multicolumn{2}{c}{$p{=}1.0$} \\
\cmidrule(lr){2-3} \cmidrule(lr){4-5} \cmidrule(lr){6-7} \cmidrule(lr){8-9}
Method & Acc. & GMACs & Acc. & GMACs & Acc. & GMACs & Acc. & GMACs \\
\midrule
\method-B (4 sub.) & 75.4 & 1.1 & 81.6 & 4.0 & 82.8 & 8.8 & 82.8 & 15.4 \\
\method-B (3 sub.) & 75.9 & 1.1 & 81.8 & 4.0 & -- & -- & 82.5 & 15.4 \\
\method-S (4 sub.) & 70.5 & 0.6 & 79.9 & 2.3 & 81.9 & 5.0 & 82.2 & 8.7 \\
\method-S (3 sub.) & 71.7 & 0.6 & 80.2 & 2.3 & -- & -- & 82.3 & 8.7 \\
\midrule
MatFormer~\cite{devvrit2024matformer} & 80.5 & 9.2 & 81.8 & 12.0 & 82.0 & 14.8 & 82.0 & 17.6 \\
DynaBERT~\cite{hou_dynabert_2020} & 73.0 & 3.4 & 80.2 & 7.5 & 81.2 & 12.2 & 81.3 & 17.6 \\
SortedNet~\cite{valipour2023sortednet} & 70.2 & 1.3 & 78.9 & 4.6 & 80.6 & 10.0 & 80.8 & 17.6 \\
HydraViT~\cite{haberer_hydravit_2024} & 70.6 & 1.3 & 79.3 & 4.6 & 81.0 & 10.0 & 81.1 & 17.6 \\
HydraViT (1100 epochs)~\cite{haberer_hydravit_2024} & 71.7 & 1.3 & 80.2 & 4.6 & 81.5 & 10.0 & 81.6 & 17.6 \\
\bottomrule
\end{tabular}
\end{table*}

We evaluate \method on ImageNet-1k across three model sizes (ConvNeXt-T, ConvNeXt-S, ConvNeXt-B) and compare against both ViT-based and classical CNN slimming methods. We describe the setup (Section~\ref{sec:setup}), present main results (Section~\ref{sec:main_results}), analyze scaling and CNN baselines (Section~\ref{sec:scaling}), evaluate AutoSlim and training duration (Section~\ref{sec:autoslim_results}), and discuss limitations (Section~\ref{sec:discussion}).

\subsection{Setup}
\label{sec:setup}

We evaluate \method on ImageNet-1k~\cite{deng_imagenet_2009,ILSVRC15}, which contains approximately 1.28M training and 50k validation images across 1,000 classes at a resolution of $224 \times 224$. We implement on top of timm~\cite{rw2019timm} and train all models from scratch for 600 epochs using AdamW~\cite{loshchilov_decoupled_2019} with a learning rate of $4 \times 10^{-3}$, weight decay of 0.05, and a cosine learning rate schedule with 20 warmup epochs. The effective batch size is 2048, achieved via gradient accumulation. We apply RandAugment~\cite{cubuk_randaugment_2020}, Mixup~\cite{zhang_mixup_2018} ($\alpha=0.8$), CutMix~\cite{yun_cutmix_2019} ($\alpha=1.0$), Random Erasing~\cite{zhong_random_2020} (probability 0.25), and label smoothing~\cite{szegedy_rethinking_2016} ($\epsilon=0.1$). We use stochastic depth~\cite{huang_deep_2016} with a drop path rate of 0.1 for ConvNeXt-T and EMA with a decay of 0.9999. Training is performed on up to 4 NVIDIA L40S GPUs (48~GB each). This follows the training procedure of~\citet{liu_convnet_2022} but without pre-training or knowledge distillation.

\begin{figure}[t]
\centering
\includegraphics[width=\linewidth]{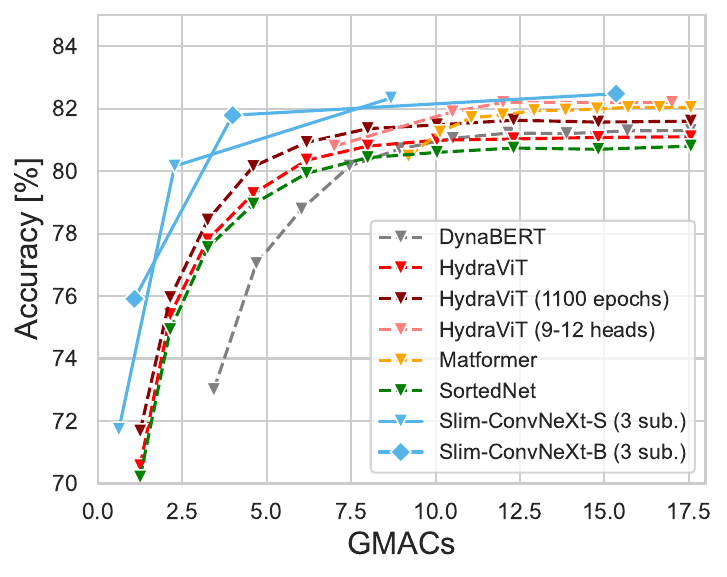}
\caption{\textbf{Accuracy vs.\ GMACs for all \method variants and ViT-based baselines on ImageNet-1k.} \method models across all three sizes achieve competitive or superior accuracy at lower GMACs compared to the ViT-based methods.}
\label{fig:baselines}
\end{figure}
We compare against HydraViT~\cite{haberer_hydravit_2024}, MatFormer~\cite{devvrit2024matformer}, SortedNet~\cite{valipour2023sortednet}, and DynaBERT~\cite{hou_dynabert_2020}, all based on either ViT-S with 4 to 6 subnetworks or ViT-B~\cite{dosovitskiy_image_2021} with 10 subnetworks each. All ViT-based baselines are finetuned from a pretrained DeiT-tiny checkpoint (300 epochs) for 300 or 800 additional epochs. We also compare against the CNN-based baselines US-Nets~\cite{yu_universally_2019} and AutoSlim~\cite{yu_autoslim_2019}. All methods are evaluated on the ImageNet-1k validation set using single-crop top-1 accuracy. GMACs are computed for a single $224 \times 224$ input image.
\subsection{Main Results}
\label{sec:main_results}

Table~\ref{tab:main_results} compares \method-S and \method-B against the ViT-B baselines. Figure~\ref{fig:baselines} visualizes the accuracy-compute trade-off across all model sizes and baselines.

Our strongest configuration, \method-B with 4 subnetworks, achieves 82.8\% top-1 accuracy at full capacity and 81.6\% at half capacity with only 4.0~GMACs. Compared to HydraViT's 10-subnetwork ViT-B finetuned for 800 epochs from a pretrained DeiT-tiny checkpoint (1100 total training epochs), this is 1.2 percentage points higher at full capacity (12 heads) and 1.4 percentage points higher at half capacity (6 heads), despite our model being trained from scratch for only 600 epochs. At quarter capacity, \method-B reaches 75.4\% at 1.1~GMACs, outperforming HydraViT's 71.7\% at a similar amount of GMACs by 3.7 percentage points. Notably, the 4-subnetwork model also achieves 82.8\% at $p{=}0.75$ with 8.8~GMACs, matching its full-capacity accuracy at roughly half the compute. This suggests that mild slimming can act as an implicit regularizer at larger model scales, an effect we do not observe for the smaller ConvNeXt-T.

\method-S with 3 subnetworks reaches 82.3\% at full capacity and 80.2\% at half capacity with 2.3~GMACs, matching HydraViT's half-capacity accuracy at roughly half the compute. With 4 subnetworks, \method-S adds a $p{=}0.75$ operating point at 81.9\% with 5.0~GMACs while retaining 82.2\% at full capacity, only 0.1 percentage points below the 3-subnetwork variant. Even \method-T with 3 subnetworks, the smallest configuration, achieves 80.8\% at full capacity and 77.4\% at half, making it competitive in the low-compute regime below 5~GMACs (see Figure~\ref{fig:hero}).

Compared to the other baselines, \method-B with 4 subnetworks outperforms DynaBERT by 1.5 and SortedNet by 2.0 percentage points at full capacity, with even larger differences at reduced widths. MatFormer achieves 82.0\% at full capacity but only flexes the MLP hidden dimension, leaving the attention layers unchanged, so its GMACs range is limited to 9.2-17.6~GMACs (roughly half to full compute). In contrast, a single \method-B covers 1.1 to 15.4~GMACs, a much wider range from a single set of weights.

An important advantage of \method is compute efficiency: because ConvNeXt avoids self-attention, our subnetworks require fewer GMACs at comparable accuracy. For instance, \method-B at half capacity achieves 81.6\% with 4.0~GMACs, surpassing HydraViT's 80.2\% at 4.6~GMACs (13\% fewer operations).

\subsection{Scaling Analysis}
\label{sec:scaling}

\begin{figure*}[t]
\centering
\includegraphics[width=\linewidth]{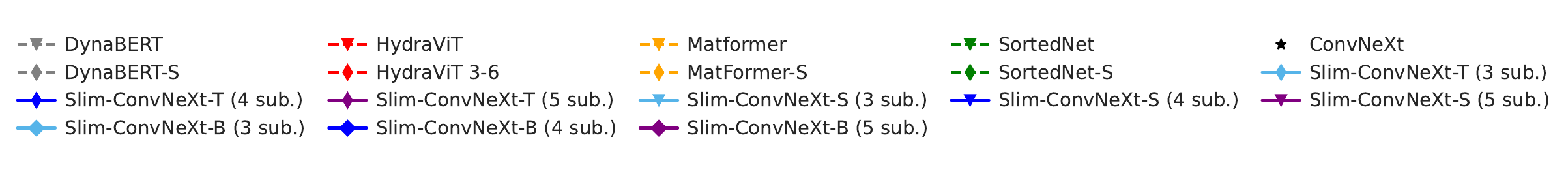}\\
\begin{subfigure}{0.32\textwidth}
\includegraphics[width=\linewidth]{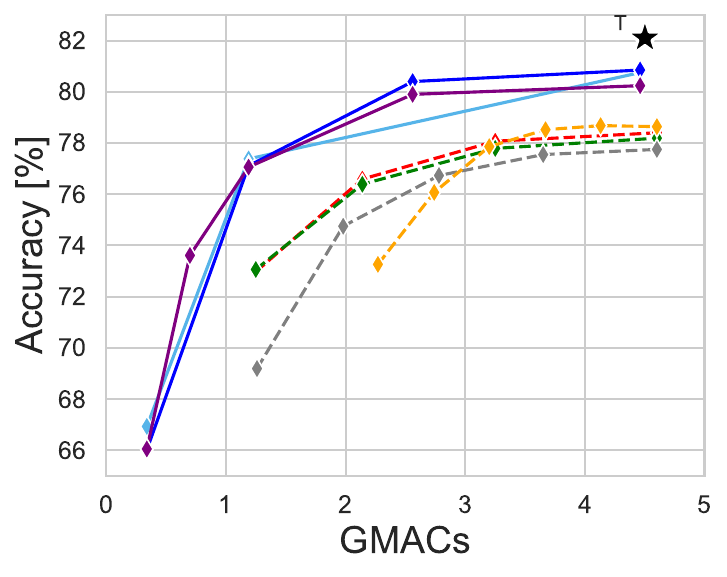}
\caption{ConvNeXt-T}
\label{fig:scaling_tiny}
\end{subfigure}
\hfill
\begin{subfigure}{0.32\textwidth}
\includegraphics[width=\linewidth]{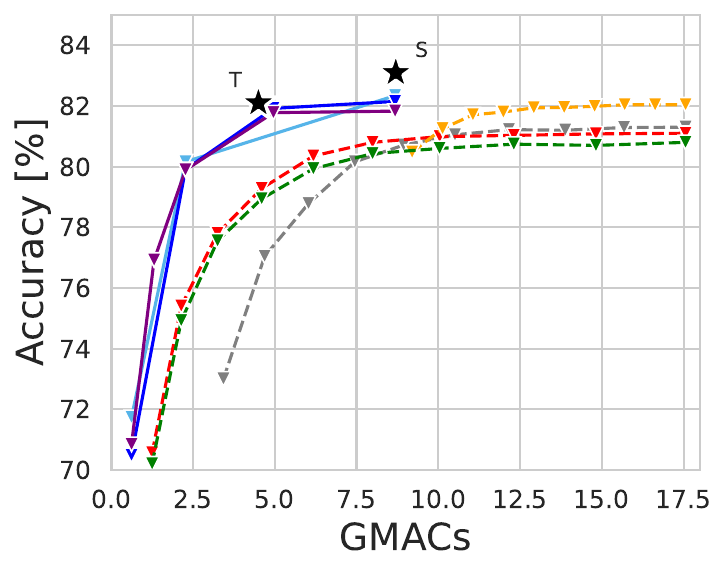}
\caption{ConvNeXt-S}
\label{fig:scaling_small}
\end{subfigure}
\hfill
\begin{subfigure}{0.32\textwidth}
\includegraphics[width=\linewidth]{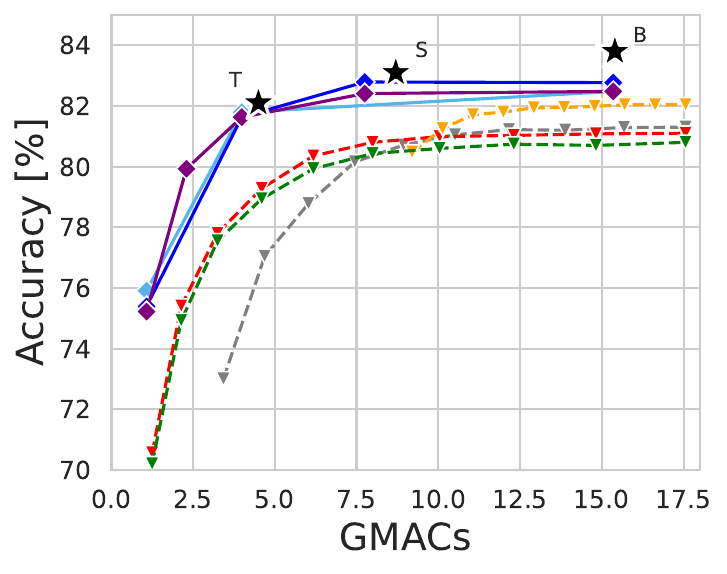}
\caption{ConvNeXt-B}
\label{fig:scaling_base}
\end{subfigure}
\caption{\textbf{Scaling comparison of \method across three ConvNeXt sizes.} Each panel shows the accuracy-compute trade-off for a single model size with its subnetwork variants, compared to ViT-based baselines. }
\label{fig:scaling}
\end{figure*}

Figure~\ref{fig:scaling} compares all three ConvNeXt sizes alongside the ViT-based baselines. For ConvNeXt-T (Figure~\ref{fig:scaling_tiny}), we compare against variants of ViT-S based baselines: HydraViT (3-6 heads), DynaBERT-S, SortedNet-S, and MatFormer-S. At 4.6~GMACs, \method-T with 3 subnetworks reaches 80.8\%, outperforming MatFormer-S (78.6\%), HydraViT (78.4\%), SortedNet-S (78.2\%), and DynaBERT-S (77.8\%) by 2.0-3.0 percentage points. At 1.2~GMACs, \method-T achieves 77.4\% compared to HydraViT's 73.0\% and SortedNet's 73.1\% at 1.3~GMACs. For ConvNeXt-S and ConvNeXt-B (Figures~\ref{fig:scaling_small} and~\ref{fig:scaling_base}), we compare against the larger ViT-B based baselines, where \method similarly achieves competitive or superior accuracy at lower GMACs. The standalone ConvNeXt models trained without slimming are shown as stars in each panel for reference.

Larger models benefit substantially more from slimming: \method-B with 4 subnetworks retains 81.6\% at half capacity, only 1.1 percentage points below its full-capacity accuracy, \method-S with 3 subnetworks drops by 2.1 percentage points from 82.3\% to 80.2\%, and \method-T with 3 subnetworks drops by 3.4 percentage points from 80.8\% to 77.4\%. This suggests that the wider channel dimensions of ConvNeXt-B provide sufficient redundancy for the first channels to carry meaningful representations even when the latter channels are removed. A similar pattern has been observed for EfficientNet~\cite{tan_efficientnet_2019}, where larger models exhibit more graceful degradation under pruning.

To quantify the cost of multi-width training, we compare the subnetworks of \method-B with 4 subnetworks against standalone ConvNeXt models trained without slimming for 300 epochs each (shown as stars in Figure~\ref{fig:scaling}). The half-capacity subnetwork (81.6\% at 4.0~GMACs) nearly matches a standalone ConvNeXt-T (82.1\% at 4.5~GMACs), and the three-quarter subnetwork (82.8\% at 8.8~GMACs) nearly matches a standalone ConvNeXt-S (83.1\% at 8.7~GMACs). At full capacity, the gap to a standalone ConvNeXt-B (83.8\%) is 1.0 percentage points. Covering this compute range with standalone models would require training all three models separately (900 total epochs and three deployments), while a single \method-B trains for 600 epochs and even provides an additional subnetwork from a single set of weights.

We also study the effect of the number of subnetworks across all three model sizes. For ConvNeXt-T, adding a fourth subnetwork at $p{=}0.75$ yields a comparable 80.9\% (vs.\ 80.8\% with 3) and fills the gap between half and full capacity, but extending to 5 subnetworks reduces full-capacity accuracy to 80.2\%. ConvNeXt-S shows a similar pattern: 4 subnetworks retain 82.2\% (vs.\ 82.3\% with 3), while 5 subnetworks drop to 81.8\%. ConvNeXt-B, however, is more robust: 4 subnetworks actually improve full-capacity accuracy to 82.8\% (vs.\ 82.5\% with 3), and even 5 subnetworks retain 82.5\%, matching the 3-subnetwork baseline. This robustness is consistent with the observation above that wider channel dimensions provide sufficient redundancy to absorb the cost of additional subnetworks. For the smaller models, each additional subnetwork shares more of the training budget, leaving fewer gradient updates per configuration. This is consistent with HydraViT~\cite{haberer_hydravit_2024}, which also observes diminishing returns when training with an increasing number of subnetworks.

\begin{figure}[t]
\centering
\includegraphics[width=\linewidth]{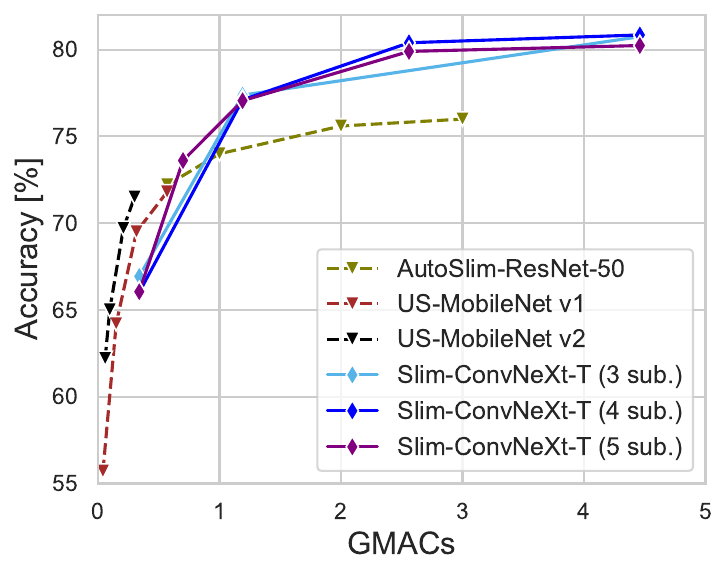}
\caption{\textbf{\method-T vs.\ classical CNN slimming methods.} \method-T with 3, 4, and 5 subnetworks compared to AutoSlim-ResNet-50~\cite{yu_autoslim_2019} and US-MobileNet v1/v2~\cite{yu_universally_2019}. \method achieves a higher accuracy ceiling and covers a wider GMACs range than the prior CNN-based approaches, whose base architectures limit their scalability.}
\label{fig:cnn_compare}
\end{figure}

Figure~\ref{fig:cnn_compare} compares \method-T against classical CNN slimming methods: AutoSlim-ResNet-50~\cite{yu_autoslim_2019} and US-MobileNet v1/v2~\cite{yu_universally_2019}. The MobileNet-based methods are limited to a narrow GMACs range below 0.6~GMACs due to their lightweight base architectures, and AutoSlim-ResNet-50 reaches at most 76.0\% at 3.0~GMACs. In contrast, \method-T covers 0.3 to 4.5~GMACs and reaches 80.8\% at full capacity, demonstrating how ConvNeXt's higher baseline accuracy and wider channel dimensions enable a much broader and more accurate scalability range for width slimming. While \method does not match MobileNet-based methods in the sub-0.6~GMACs regime, modern mobile SoCs can comfortably run models with up to 5~GMACs in real time, for instance, EfficientFormer-L3 (3.9~GMACs) achieves 3.0~ms latency on an iPhone~12~\cite{li2022efficientformer}. This makes the higher compute budget of ConvNeXt-based slimming practical for on-device deployment, while delivering substantially higher accuracy.

\subsection{AutoSlim and Training Duration}
\label{sec:autoslim_results}

\begin{figure}[t]
\centering
\includegraphics[width=\linewidth]{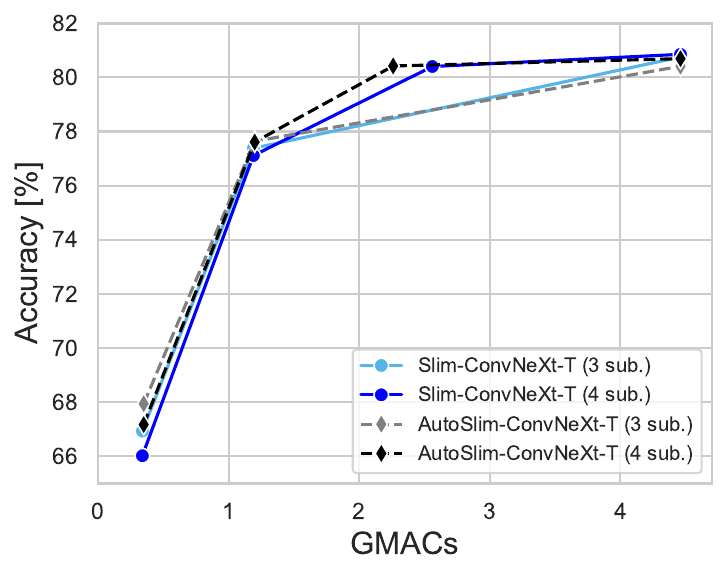}
\caption{\textbf{AutoSlim vs.\ uniform slimming on ConvNeXt-T with 3 and 4 subnetworks.} AutoSlim provides modest improvements at low capacity for both subnetwork counts, while uniform slimming retains a slight edge at full capacity.}
\label{fig:autoslim_subs}
\end{figure}

Figure~\ref{fig:autoslim_subs} compares uniform slimming with AutoSlim on ConvNeXt-T for both 3 and 4 subnetworks. AutoSlim improves quarter-capacity accuracy to 67.9\%, up from 66.9\% with uniform slimming, but at half and full capacity the benefit is marginal or slightly negative (80.4\% vs.\ 80.8\% at full capacity). With 4 subnetworks, AutoSlim shows a similar pattern.
The non-uniform p-lists found by AutoSlim preserve wider early and late blocks while slimming middle stages more aggressively, but the gains are modest, which we attribute to the zero-padding strategy: aggressively slimmed blocks pass mostly zeros through the residual path, limiting the benefit of non-uniform allocation.

\begin{figure}[t]
\centering
\includegraphics[width=\linewidth]{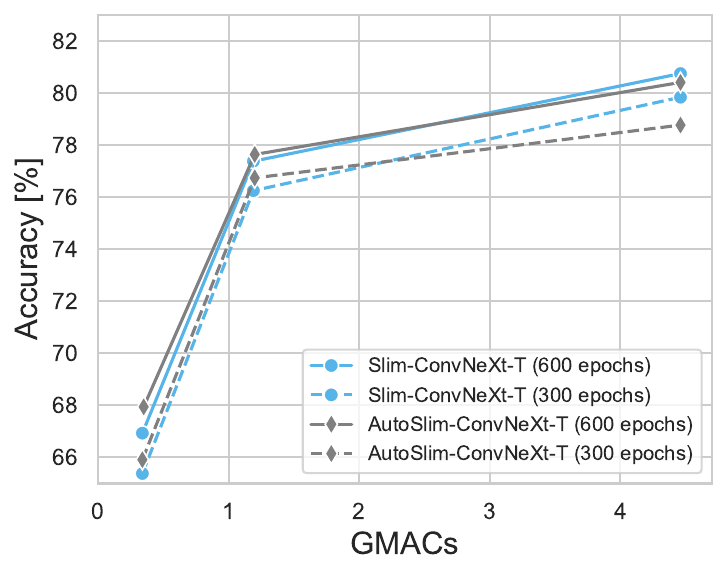}
\caption{\textbf{Training duration and AutoSlim on ConvNeXt-T (3 subnetworks).} Standard and AutoSlim models at 300 and 600 epochs. Longer training benefits all operating points, especially at low capacity.}
\label{fig:autoslim}
\end{figure}

Figure~\ref{fig:autoslim} shows the impact of training duration on ConvNeXt-T with 3 subnetworks. Extending training from 300 to 600 epochs improves accuracy for all subnetworks, with the largest gains at low capacity (65.4\% to 66.9\% at quarter capacity vs.\ 79.8\% to 80.8\% at full capacity). This suggests that low-width subnetworks, having fewer active parameters, benefit disproportionately from longer training.

\subsection{Discussion}
\label{sec:discussion}

Our results show that channel-width slimming transfers effectively to ConvNeXt, with two key architectural advantages. First, LayerNorm normalizes per sample and adapts naturally to any channel count, eliminating the switchable batch normalization required by classical slimmable CNNs~\cite{yu_slimmable_2018,yu_universally_2019}. While ViT-based baselines also use LayerNorm, their ViT backbone still carries the cost of self-attention at every operating point. Second, the inverted bottleneck structure concentrates parameters in pointwise layers that are straightforward to slice, and we believe these advantages generalize to other modern CNNs sharing similar design principles.

The main limitation is the zero-padding strategy for residual connections when applying AutoSlim: at $p{=}0.25$, up to three-quarters of the learned features can be discarded through the residual. Replacing zero-padding with a lightweight linear projection could improve accuracy at low $p$ values at the cost of additional parameters. We also note that GMACs do not capture memory access patterns or hardware-specific latencies~\cite{ma_shufflenet_2018}. Depthwise convolutions may underperform their GMAC count on throughput-oriented GPUs due to low arithmetic intensity~\cite{yu_inceptionnext_2024}, while being efficient on memory-bandwidth-limited mobile SoCs.

The asymmetry in slimming degradation across scales (Section~\ref{sec:scaling}) suggests that for deployment scenarios requiring reduced capacity, starting from a larger base model and slimming aggressively may yield better results than training a smaller model at full width.

\section{Conclusion}
\label{sec:conclusion}

We present \method, the first application of channel-width slimming to the ConvNeXt~\cite{liu_convnet_2022} architecture family. By slicing channel dimensions within each ConvNeXt block and alternating between width configurations during training, our approach induces multiple nested subnetworks within a single set of shared weights, analogous to various Vision Transformer approaches, such as HydraViT~\cite{haberer_hydravit_2024}, MatFormer~\cite{devvrit2024matformer}, SortedNet~\cite{valipour2023sortednet}, and DynaBERT~\cite{hou_dynabert_2020}, but with the simplicity and efficiency of convolutions. The resulting elastic model can dynamically adjust its computational cost at inference time, enabling deployment scenarios where a single model adapts to varying resource budgets, whether across a heterogeneous device fleet or on a single device responding to battery constraints, thermal throttling, or latency deadlines.

On ImageNet-1k~\cite{deng_imagenet_2009}, \method-T with 3 subnetworks achieves 80.8\% top-1 accuracy at 4.5~GMACs and 77.4\% at 1.2~GMACs, exceeding HydraViT's 6-head subnetwork (78.4\% at 4.6~GMACs) by 2.4 percentage points and its 3-head configuration (73.0\%) by 4.4 percentage points. Scaling to \method-B further improves accuracy to 82.8\% at 15.35~GMACS, with larger models retaining higher accuracy under slimming. We find that 4 subnetworks strike the best balance between flexibility and peak accuracy, and that AutoSlim~\cite{yu_autoslim_2019} offers modest gains at small capacities by allocating more width to early and late blocks.

\section*{Acknowledgements}

This research received funding from the Federal Ministry for Economic Affairs and Energy under the CAPTN X-FERRY project (grant no.~FK: 03SX612A). It was supported in part by high-performance computing resources provided by the Kiel University Computing Centre and the Hydra computing cluster, funded by the German Research Foundation (grant no.~442268015) and the Petersen Foundation (grant no.~602157), respectively.

{
    \small
    \bibliographystyle{ieeenat_fullname}
    \bibliography{main}
}


\end{document}